\documentclass{article}
\usepackage{graphicx}
\usepackage{amsmath, amssymb}
\usepackage{array}
\usepackage{booktabs}
\usepackage{geometry}
\usepackage{adjustbox}
\usepackage{longtable}
\usepackage{tikz}
\usepackage{hyperref}
\usepackage{algorithm}
\usepackage{algorithmic}
\usepackage{amssymb}
\usepackage{tcolorbox}
\usepackage{palatino}
\usepackage{colortbl}

\begin{document}

\title{Introduction to Reinforcement Learning}
\author{
    $\text{Majid Ghasemi}^{\star, 1}$ and $\text{Dariush~Ebrahimi}^{1}$ \\
    {\small $^1$Department of Computer Science, Wilfrid Laurier University, Waterloo, Canada} \\
    {\small \texttt{\{mghasemi$^\star$, debrahimi\}@wlu.ca}} \\
    {\small $^\star$Corresponding author} 
}

\date{}

\maketitle

\begin{abstract}
Reinforcement Learning (RL), a subfield of Artificial Intelligence (AI), focuses on training agents to make decisions by interacting with their environment to maximize cumulative rewards. This paper provides an overview of RL, covering its core concepts, methodologies, and resources for further learning. It offers a thorough explanation of fundamental components such as states, actions, policies, and reward signals, ensuring readers develop a solid foundational understanding. Additionally, the paper presents a variety of RL algorithms, categorized based on the key factors such as model-free, model-based, value-based, policy-based, and other key factors. Resources for learning and implementing RL, such as books, courses, and online communities are also provided. By offering a clear, structured introduction, this paper aims to simplify the complexities of RL for beginners, providing a straightforward pathway to understanding and applying real-time techniques.
\end{abstract}

\section{Introduction}
Reinforcement Learning (RL) is a subfield of Artificial Intelligence (AI) that focuses on training by interacting with the environment, aiming to maximize cumulative reward over time \cite{sutton2018reinforcement}. In contrast to supervised learning, where the objective is to learn from labeled examples, or unsupervised learning, which is based on detecting patterns in the data, RL deals with an autonomous agent that must make intuitive decisions and consequently learn from its actions, often without existing data. The key idea is to learn how the world works (e.g., what action gets a (positive) reward and which does not) to maximize cumulative rewards over time through trial-and-error exploration. RL revolves around several key concepts: States, Actions, Policy, Rewards, Transition Dynamics(Probabilities), and Environment Model. Each of these components plays a crucial role in defining the agent's interaction with its environment and the overall learning process, and are defined over the next paragraphs. This paper assumes readers to have basic knowledge of Machine Learning (ML) algorithms like Supervised and Unsupervised Learning. 

A \textbf{State} ($s \in \mathcal{S}$) represents a specific condition or configuration of the environment at a given time as perceived by the agent. A state sets the scene for the agent to make choices and select actions, which describes all states from which an agent can choose for each action that occurs. For example, in chess, a state might be one specific layout of pieces on the board. In brief, states are the situations that an agent can be in and observe to make decisions. States can be discrete (the given example) and continuous (e.g., the position of a robot, in terms of $x$ and $y$ coordinates).
\textbf{Actions} ($a \in \mathcal{A}$) are the set of possible moves or decisions an agent can make while interacting with the environment. A selected action is part of the strategy followed by an agent to reach its desired goals according to its current states and policy. In the chess example, moving a piece from one square to another is an action. Similar to states, actions can be discrete and continuous. The given chess example has a set of discrete actions (moving the piece according to the rules, which has a finite set of actions). For continuous actions, let us consider the robot example. A robot can change its coordinates and its movement speed, which both are continuous.
\textbf{Policy} ($\pi$) guides the behavior of a learning agent by mapping perceived states of the environment into actions. This could be a simple function, a lookup table, or a complex computation, e.g., approximated by Neural Networks. Policies can be stochastic, defining the likelihood of taking certain actions. A \textit{stochastic policy} ($\pi(a|s)$) defines a probability distribution over possible actions for a given state. Instead of selecting a single action, it samples actions based on the probabilities defined by the policy.
On the other hand, A \textit{deterministic policy} ($\pi(s) = a$) directly maps a state to a specific action. For a given state, the action taken is fixed and does not involve any randomness. In the game of chess, an agent's policy determines the move it makes given the current board configuration (state). Imagine an agent evaluating its options in a given state of the board. It assigns probabilities to each legal move based on their estimated utility. For instance: $\pi(a_1 | s) = 0.5$, $\pi(a_2 | s) = 0.3$, $\pi(a_3 | s) = 0.2$.
Here, $a_1$, $a_2$, and $a_3$ are possible moves, and the policy specifies a 50\% chance of selecting $a_1$, 30\% for $a_2$, and 20\% for $a_3$. The agent chooses its move by sampling from this distribution. This approach introduces diversity in decision-making, which can be particularly useful when exploring novel strategies or dealing with uncertain opponents. Conversely, a deterministic policy maps the current state $s$ to a specific move $a$ (e.g., $\pi(s) = a_1$).  
In this case, the agent always chooses $a_1$ (e.g., moving the queen to a specific square) when the board configuration matches $s$. 
Deterministic policies are efficient in well-defined and predictable scenarios, such as when the agent has already learned the optimal moves for most board configurations. \textbf{Rewards} ($r \in \mathcal{R}$) are a critical factor in RL since it provides the agent with an objective at each time step, defining both local and global goals that the agent aims to achieve over time. Rewards differentiate positive from negative events and help update policies according to the outcomes of actions. Rewards depend on the state of the environment and the actions taken. For instance, in a Chess scenario, winning the game will lead to a positive reward, while losing points will lead to a negative reward. Also, the game might end up in a draw, which lead to no reward. Reward function defines the immediate reward obtained when taking action $a$ in state $s$. It can be deterministic ($\mathcal{R}(s,r) = r$) or stochastic ($P(r|s,a)$). \textbf{Transition Dynamics} function defines the probability of reaching a new state $s'$ given the current state $s$ and action $a$. It can be written mathematically as $\mathcal{P}(s'|s, a)$. \textbf{Environment Model} is an approximation or prediction of the environment, mapping out what will be available (the next state and reward) given a particular input from the state and action. These models help with planning by identifying what actions should be taken based on possible future events. RL approaches that use models are called \textit{model-based} methods, whereas those relying solely on trial-and-error learning are \textit{model-free} methods. This characteristic is an important factor to choose the appropriate algorithms, suitable for the problem. The environment's dynamics can be represented as a transition function and a reward function. Together, these form the Markov Decision Process (MDP):

\begin{equation}
    \nonumber
    \mathcal{M} = (\mathcal{S}, \mathcal{A}, \mathcal{P}, \mathcal{R}, \gamma)
\end{equation}
where $\mathcal{S}$ is state space, $\mathcal{A}$ is action space, $\mathcal{P}(s'|s, a)$ is the transition dynamics, $\mathcal{R}(s,a)$ is the Reward function, and $\gamma \in \left[ 0, 1 \right]$. $\gamma$ is the discount factor, a parameter that determines how much importance is given to future rewards compared to immediate rewards.

Figure 1 illustrates the interaction between an RL agent and its environment. The agent observes the current state ($\mathcal{S}_t$), selects an action ($\mathcal{A}_t$) based on its policy ($\pi$), and receives a reward based on the reward distribution ($\mathcal{R}_t$) along with the next state ($\mathcal{S}_{t+1}$). This feedback loop is critical for the agent to learn and update its policy to maximize cumulative rewards.

\begin{figure}[h]
\centering
\includegraphics[width=0.8\textwidth]{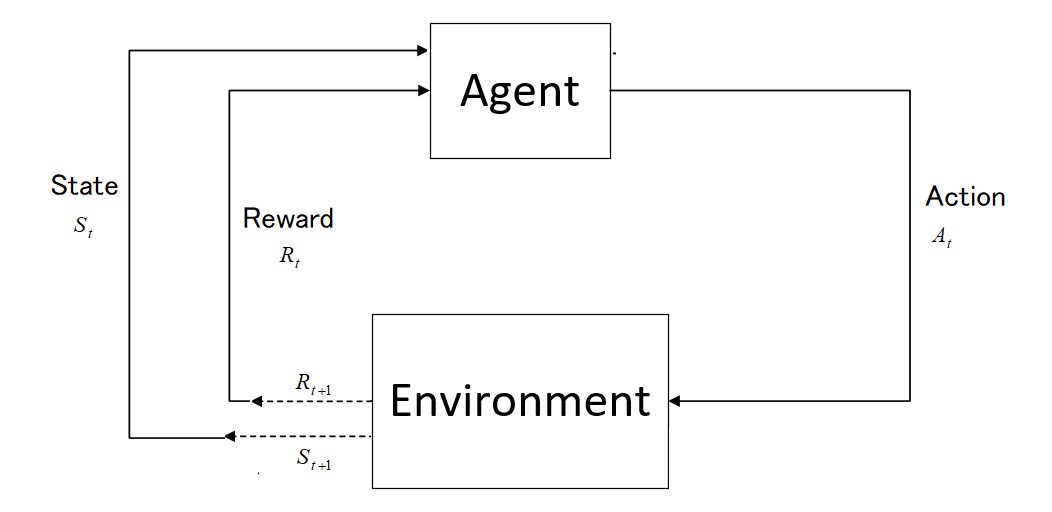}
\caption{Overview of Reinforcement Learning \cite{sutton2018reinforcement}}
\label{fig:overview_rl}
\end{figure}


The rest of the paper is organized as follows: Section~\ref{sec:multi-bandits} introduces backgrounds and ket concepts in RL, starting with multi-armed bandit. Bandits are a great place to start learning RL's foundational concepts like value functions and the Bellman equations. A complete introduction to core RL methods is given in section~\ref{sec:Terminology}. Section~\ref{sec:Algs} analyzes essential RL algorithms, categorized them in a comprehensive manner. In section~\ref{sec:Resources}, useful resources for learning RL are provided for readers would like to delve deeper in the realm of RL. Finally, section~\ref{sec:conclusion} Concludes the paper.

\section{Backgrounds \& Key Concepts}\label{sec:multi-bandits}



Understanding RL requires a solid grasp of its foundational principles and the mathematical frameworks that guide agent-environment interactions. This section introduces the essential building blocks of it, starting with the simplest form of decision-making under uncertainty: the Multi-Armed Bandit problem. This provides an intuitive gateway to understanding how agents learn from evaluative feedback.
Building on this, we delve into MDPs, which formalize sequential decision-making by balancing immediate and future rewards. Finally, we discuss key RL metrics, such as value functions and policies, which form the backbone of most RL algorithms. Together, these topics lay the groundwork for exploring the core RL methods in the next section.

\subsection{Multi-Armed Bandit(s)}
The Multi-Armed Bandit problem serves as a foundational example for understanding Reinforcement Learning (RL). It represents a simplified decision-making scenario where an agent repeatedly chooses from $\mathcal{K}$ actions (or arms) to maximize cumulative rewards over time. Each action is associated with an unknown reward distribution, requiring the agent to balance \textit{exploration} (gathering information about all actions) and \textit{exploitation} (maximizing immediate rewards using the best-known action).
Unlike supervised learning, where instructive feedback explicitly points to correct actions, RL often involves evaluative feedback. This type of feedback only signals the quality of chosen actions, making it harder to identify optimal strategies without extensive interaction with the environment.


In the $\mathcal{K}$-armed bandit problem, each action \( a \in \{a_1, a_2, \dots, a_k\} \) yields a reward drawn from a stationary probability distribution. The \textit{stationarity} assumption implies that the reward probabilities remain constant over time \cite{li2022testing}. For each of the $k \in \mathcal{K}$ actions available, there is an expected average reward, referred to as the value of the action \cite{sutton2018reinforcement}. Let $\mathcal{A}_t$ denote the action chosen at time step \( t \), and $\mathcal{R}_t$ the corresponding reward received. The true value of an action, \( q_*(a) \), is defined as the expected reward by taking that action as:

\begin{equation}
    q_*(a) = \mathbb{E}[\mathcal{R}_t \mid \mathcal{A}_t = a]
\end{equation}
If \( q_*(a) \) were known for all \( a \), the problem would be trivial: the agent would always choose the action with the highest \( q_*(a) \). However, in practice, these values are unknown and must be estimated through interaction with the environment.
After estimating action values, there is at least one action with the highest estimated value at each time step. These are called greedy actions. Selecting a greedy action exploits current knowledge for immediate reward, while selecting non-greedy actions explores to improve estimates. Balancing exploration and exploitation is crucial in RL. Opting for exploitation maximizes immediate reward, while exploration can yield higher overall reward depending on various factors \cite{sutton2018reinforcement, thrun1992efficient}.
Estimating action values through sampling averages is efficient for stationary bandit problems. In real-world problems with non-stationary environments, it makes sense to weight recent rewards more. Using a constant step-size parameter is popular. The following update rule can be used to update the action value, which is known as \textbf{sample-average method}.

\begin{equation}
    \label{Eq_AIA10}
    \mathcal{Q}_t(a) = \frac{\sum_{i=1}^{t-1} \mathcal{R}_i \cdot \mathbf{1}_{\{\mathcal{A}_i = a\}}}{\sum_{i=1}^{t-1} \mathbf{1}_{\{\mathcal{A}_i = a\}}},
\end{equation}
where \( \mathbf{1}_{\{\mathcal{A}_i = a\}} \) is an indicator function that equals 1 if action \( a \) was chosen at step \( i \), and 0 otherwise. Over time, \( \mathcal{Q}_t(a) \) converges to \( q_*(a) \) by the law of large numbers \cite{erd1970new}, provided all actions are sampled sufficiently often.

\begin{equation}
\mathbf{1}_{\{\mathcal{A}_i = a\}} = \begin{cases} 
1 & \texttt{if action } a \texttt{ is taken at time step } i \\
0 & \texttt{otherwise} 
\end{cases}
\end{equation}
Equation~\ref{Eq_AIA10} can be written in a different way as follows, which represents the action value \( \mathcal{Q}_t(a) \) as the average of the rewards received from that action up to time step \( t \).
\begin{equation}
\mathcal{Q}_t(a) = \frac{\texttt{sum of rewards when the action } a \texttt{ is taken prior to time step } t}{\texttt{number of times the action } a \texttt{ is taken prior to time step } t}
\end{equation}

\begin{equation}
    \label{Eq_Qn}
    \mathcal{Q}_{n+1} = \mathcal{Q}_n + \alpha [\mathcal{R}_n - \mathcal{Q}_n],
\end{equation}
To iteratively refine \( \mathcal{Q}_t(a) \), we use Equation~\ref{Eq_Qn}, known as \textbf{incremental update rule}, where \( \mathcal{Q}_{n+1} \) is the updated action value after observing reward $\mathcal{R}_n$. $\mathcal{Q}_n$ is the previous estimate, and $\alpha$ is the learning rate (a step-size parameter determining how much new information overrides the old estimate).
This update rule incrementally adjusts the action value based on the difference between the received reward and the current estimate, weighted by the learning rate \( \alpha \). This iterative process allows the agent to refine its value estimates and improve decision-making over time. In other words, $\texttt{NewEstimate} \leftarrow \texttt{OldEstimate} + \texttt{StepSize} [\texttt{Target} - \texttt{OldEstimate}]$ \cite{sutton2018reinforcement}.

One of the critical challenges in solving the multi-armed bandit problem, and in RL as well, is balancing exploration and exploitation, the greedy action selection ($\mathcal{A}_t = \arg\max_a \mathcal{Q}_t(a)$). Several strategies address this trade-off. We touch some of the widely used ones here, starting with \textbf{$\epsilon-$greedy}. In this method, with probability \( \epsilon \), choose a random action (exploration); otherwise, select the greedy action (exploitation). This ensures that all actions are sampled, albeit less frequently for suboptimal actions. Another method, known as \textbf{Upper Confidence Bound (UCB)}, which balances exploitation and exploration by incorporating uncertainty into action selection \cite{garivier2011upper}: 
\begin{equation}
        \mathcal{A}_t = \arg\max_a \left[ \mathcal{Q}_t(a) + c \sqrt{\frac{\ln t}{\mathcal{N}_t(a)}} \right],
\end{equation}
where \( \mathcal{N}_t(a) \) is the number of times action \( a \) has been chosen, and \( c > 0 \) controls the exploration rate. UCB prioritizes actions with high uncertainty or fewer samples.
\textbf{Optimistic Initialization}, assigns high initial values to \( \mathcal{Q}_1(a) \), encouraging exploration of all actions early on. For example, setting \( \mathcal{Q}_1(a) = +5 \) makes untried actions appear attractive initially.

These methods ensure all actions are sampled enough to estimate their true value accurately, making the selection of the best action almost certain over time. However, these are theoretical long-term benefits and may not directly indicate practical effectiveness \cite{ladosz2022exploration}. To begin with, estimating action values by sampling averages is inefficient for large numbers of samples ($\mathcal{K} > n$). A more efficient approach derives $\mathcal{Q}_n$ as:
\begin{equation}
\mathcal{Q}_n \equiv \frac{\mathcal{R}_1 + \mathcal{R}_2 + \cdots + \mathcal{R}_{n-1}}{n-1}
\end{equation}
We can express the update rule for action value recursively
\begin{equation}
\mathcal{Q}_{n+1} = \mathcal{Q}_n + \frac{1}{n} [\mathcal{R}_n - \mathcal{Q}_n]
\end{equation}
This recursive equation requires memory only for $\mathcal{Q}_n$ and $n$, with minimal computation for each reward \cite{luo2024survey}.
Sample averages apply to stationary bandit problems. In non-stationary environments, recent rewards are more relevant. Using a decaying constant step-size parameter, issues related to non-stationary problems can be mitigated \cite{sutton2018reinforcement}.

\begin{algorithm}
	\caption{A simple bandit algorithm}
	\begin{algorithmic}[1]
		\STATE \textbf{Initialize:}
		\FOR{$a = 1$ to $k$}
		\STATE $\mathcal{Q}(a) \gets 0$
		\STATE $\mathcal{N}(a) \gets 0$
		\ENDFOR
		\STATE \textbf{Repeat forever:}
		\STATE $\mathcal{A} \gets 
		\begin{cases} 
			\arg\max_a \mathcal{Q}(a) & \text{with probability } 1 - \epsilon \\ 
			\text{a random action} & \text{with probability } \epsilon 
		\end{cases}$
		\STATE $R \gets \texttt{bandit}(\mathcal{A})$
		\STATE $\mathcal{N}(\mathcal{A}) \gets \mathcal{N}(\mathcal{A}) + 1$
		\STATE $\mathcal{Q}(\mathcal{A}) \gets \mathcal{Q}(\mathcal{A}) + \frac{1}{\mathcal{N}(\mathcal{A})} \big[\mathcal{R} - \mathcal{Q}(\mathcal{A})\big]$
	\end{algorithmic}
\end{algorithm}

The initial estimates of action values, \( \mathcal{Q}_1(a) \), play a crucial role in the learning process. These initial guesses influence the early decisions made by the agent. While sample-average methods can reduce this initial bias after each action is chosen at least once, methods using a constant step-size parameter, \( \alpha \), tend to mitigate this bias more gradually over time. Setting optimistic initial values can be particularly advantageous. By assigning higher initial estimates (e.g., +5), the agent is encouraged to explore more actions early on. This is because the initial optimism makes untried actions appear more attractive, thus promoting exploration even when the agent uses a greedy strategy. This approach helps ensure that the agent thoroughly investigates the action space before converging to a final policy. However, this strategy requires careful consideration in defining the initial values, which are often set to zero in standard practice. The choice of initial values should reflect an informed guess about the potential rewards, and overly optimistic values can prevent the agent from converging efficiently if not properly managed.
Overall, optimistic initial values can be a useful technique to balance exploration and exploitation in RL, encouraging broader exploration and potentially leading to more optimal long-term policies \cite{wiering2012reinforcement}.

Over the next subsection, another fundamental concept in RL, MDPs, are introduced.

\subsection{Markov Decision Process (MDPs)}\label{sec:FMDP}
An MDP provides a framework for sequential decision-making in which actions affect immediate rewards as well as future outcomes. In MDPs, immediate rewards are balanced with delayed rewards. In contrast to bandit problems in which the goal is to determine the value of each action $a$, MDPs aim to measure the value of taking action $a$ in state $s$, or the value of being in state $s$ assuming optimal actions are taken. A correct assessment of the long-term effects of interventions requires the estimation of these state-specific values \cite{sutton2018reinforcement}. MDPs consist of states, actions, and rewards $(\mathcal{S, A, R})$. Discrete probability distributions are assigned to the random variables $\mathcal{R}_t$ and $\mathcal{S}_t$ based on the preceding state and action. Using the probabilities of occurrence of the random variables $\mathcal{R}_t$ and $\mathcal{S}_t$, derive equations for these variables. A system is considered Markovian when the outcome of an action is independent of past actions and states, relying solely on the current state \cite{van2012reinforcement}. The Markov property requires the state to encapsulate significant details of the entire past interaction influencing future outcomes \cite{kaelbling1996reinforcement}. This definition is the basis of MDPs being used in RL.
To describe the dynamics of an MDP, we use the state-transition probability function $p(s', r \mid s, a)$, which is defined as follows:

\begin{equation}
p(s', r \mid s, a) \equiv \Pr\{\mathcal{S}_t = s', \mathcal{R}_t = r \mid \mathcal{S}_{t-1} = s, \mathcal{A}_{t-1} = a\}
\end{equation}
where the function $p$ defines the MDP dynamics.
The following state-transition probabilities, state-action and state-action-next-state triple rewards can be derived from the four-argument dynamic function $p$. We can derive the state-transition probabilities, the expected reward for state-action pairs, and the expected rewards for state-action-next-state triples as follows:

\begin{equation}
p(s' \mid s, a) \equiv \Pr\{\mathcal{S}_t = s' \mid \mathcal{S}_{t-1} = s, \mathcal{A}_{t-1} = a\} = \sum_{r \in \mathcal{R}} p(s', r \mid s, a)
\end{equation}
\begin{equation}
r(s, a) \equiv \mathbb{E}\{\mathcal{R}_t \mid \mathcal{S}_{t-1} = s, \mathcal{A}_{t-1} = a\} = \sum_{r \in \mathcal{R}} r \sum_{r \in \mathcal{R}} p(s', r \mid s, a)
\end{equation}
\begin{equation}
r(s, a, s') \equiv \mathbb{E}\{\mathcal{R}_t \mid \mathcal{S}_{t-1} = s, \mathcal{A}_{t-1} = a, \mathcal{S}_t = s'\} = \sum_{r \in \mathcal{R}} \frac{r p(s', r \mid s, a)}{p(s' \mid s, a)}
\end{equation}

The concept of actions encompasses any decisions relating to learning, and the concept of states encompasses any information that is available in order to inform those decisions. As part of the MDP framework, goal-directed behavior is abstracted through interaction. Any learning problem can be reduced to three signals between an agent and its environment: actions, states, and rewards. A wide range of applications have been demonstrated for this framework \cite{sutton2018reinforcement}.
We are now able to formally define and solve RL problems. We have defined rewards, objectives, probability distributions, the environment, and the agent. Some concepts, however, were defined informally. According to our statement, the agent seeks to maximize future rewards, but how can this be mathematically expressed?
The return, denoted \( G_t \), is the cumulative sum of rewards received from time step \( t \) onwards. For episodic tasks, it is defined as follows:

\begin{equation}
\mathcal{G}_t \equiv \mathcal{R}_{t+1} + \mathcal{R}_{t+2} + \ldots + \mathcal{R}_\mathcal{T}
\end{equation}

Here, $\mathcal{G}_t$ is a specific function of the reward sequence. Episodic problems are those in which the interactions between agents and their environment occur naturally in sequence, known as episodes, and tasks are termed episodic tasks. The game hangman is a good example of this. At the end of each episode, a standard starting state is restored. The term \textit{new games} refers to the next state after the terminal state, which is the final state leading to the end of an episode. It is common for ongoing tasks to involve interactions that persist continuously throughout the duration of the task, such as process control or applications that utilize robots with prolonged lifespans. The term \textit{continuing tasks} refers to these activities.
As there are no terminal states in continuing tasks $(\mathcal{T} = \infty)$, the return for continuing tasks should be defined differently. It is possible that the return could be infinite if the agent consistently receives a reward. For continuing tasks, where there is no terminal state, the return \( \mathcal{G}_t \) is defined as the discounted sum of future rewards:

\begin{equation}
\mathcal{G}_t \equiv \mathcal{R}_{t+1} + \gamma \mathcal{R}_{t+2} + \gamma^2 \mathcal{R}_{t+3} + \ldots = \sum_{k=0}^\infty \gamma^k \mathcal{R}_{t+k+1}
\end{equation}
where $\gamma$ is the discount rate $(0 \leq \gamma \leq 1)$.
The discount rate affects the current worth of future rewards. When $\gamma < 1$, the infinite sum converges to a finite value. With $\gamma = 0$, the agent maximizes immediate rewards. As $\gamma$ approaches 1, future rewards carry more weight. We can also express the return \( \mathcal{G}_t \) recursively
\begin{equation}
\mathcal{G}_t \equiv \mathcal{R}_{t+1} + \gamma \mathcal{G}_{t+1}
\end{equation}
The return is finite if the reward is non-zero and constant, and $\gamma < 1$. Equation~\ref{Eq_Gt} works for both episodic and continuing tasks if $\mathcal{T} = \infty$ or $\gamma = 1$, respectively.
\begin{equation}
\label{Eq_Gt}
\mathcal{G}_t \equiv \sum_{k=t+1}^\mathcal{T} \gamma^{k-t-1} \mathcal{R}_k
\end{equation}
The concepts introduced in this section—ranging from bandits and MDPs to value functions and policies—provide the mathematical and conceptual tools required to understand RL methods. In the next subsection, we explore how these foundational ideas evolve into core RL algorithms, bridging theory and application.

\subsection{Policies and Value Functions}\label{sec:PVF}
The value function estimates the expected return of the agent being in a certain state (or performing an action in a particular state). Depending on the actions selected, these factors will vary. There is a link between value functions and policies, which are linked to probabilities of action based on states. Value functions can be divided into two broad categories; \textbf{State Value Functions} and \textbf{Action Value Functions}. The value function of a state $s$ under policy $\pi$, $v_\pi(s)$, is the expected return starting in $s$ and following $\pi$ thereafter (Equation~\ref{Eq_VF}). On the other hand, the value of taking action $a$ in state $s$ under policy $\pi$, $q_\pi(s,a)$, is the expected return starting from $s$, taking action $a$, and following $\pi$ thereafter (Equation~\ref{Eq_QF}).
\begin{equation}
    \label{Eq_VF}
    v_\pi(s) \equiv \mathbb{E}_\pi \left[\sum_{k=0}^\infty \gamma^k \mathcal{R}_{t+k+1} \mid \mathcal{S}_t=s\right], \quad \text{for all } s \in \mathcal{S}
\end{equation}
\begin{equation}
\label{Eq_QF}
q_\pi(s,a) \equiv \mathbb{E}_\pi [\mathcal{G}_t \mid \mathcal{S}_t=s, \mathcal{A}_t=a] = \mathbb{E}_\pi \left[\sum_{k=0}^\infty \gamma^k \mathcal{R}_{t+k+1} \mid \mathcal{S}_t=s, \mathcal{A}_t=a\right]
\end{equation}
It is important to note the difference between $v$ and $q$, namely that $q$ depends on the actions taken in each state. With ten states and eight actions per state, $q$ requires 80 functions, while $v$ requires only 10 functions. Following policy $\pi$, if an agent averages returns from each state, the average converges to $v_\pi(s)$. Averaging returns from each action converges to $q_\pi(s,a)$ \cite{sutton2018reinforcement}. $v_\pi(s)$ can be written recursively:
\begin{equation}
    \label{Eq_Bellman}
    v_\pi(s) \equiv \mathbb{E}_\pi [\mathcal{G}_t \mid \mathcal{S}_t=s] = \mathbb{E}_\pi [\mathcal{R}_{t+1} + \gamma \mathcal{G}_{t+1} \mid \mathcal{S}_t=s]
= \sum_a \pi(a|s) \sum_{s'} \sum_r p(s',r|s,a) [r + \gamma v_\pi(s')]
\end{equation}
Equation~\ref{Eq_Bellman} is the Bellman equation for $v_\pi$. The Bellman equation relates the value of a state to its potential successor states' values. The diagram illustrates the anticipation from a state to its successors. The value of the initial state equals the discounted value of the expected next state plus the anticipated reward \cite{o2018uncertainty, sutton2018reinforcement}.

\( v_\pi(s) \) and \( q_\pi(s, a) \) serve different purposes in RL. 
In the evaluation of deterministic policies or when understanding the value of being in a particular state is required, \textbf{state-value functions} are used. In policy evaluation and policy iteration methods, where a policy is explicitly defined and it is necessary to evaluate the performance of being in a particular state under the policy, these methods are highly useful. The use of state-value functions is beneficial when there are many actions, since they reduce complexity by requiring only an evaluation of state values.
\textbf{Action-value functions}, on the other hand, are used to evaluate and compare the potential for different actions when they are taking place in the same state. They are crucial for the selection of actions, where the goal is to determine the most appropriate action for each situation. As action-value functions take into account the expected return of different actions, they are particularly useful in environments with stochastic policies. Moreover, when dealing with continuous action spaces, action-value functions can provide a more detailed understanding of the impact of actions, aiding in the fine-tuning of policy implementation.

\textit{Example: }Consider a gambling scenario where a player starts with \$10 and faces decisions regarding the amount to bet. This game illustrates state and action value functions in RL.
\textbf{State Value function ($v^\pi(s)$)} quantifies expected cumulative future rewards for a state $s$, given policy $\pi$. Suppose the player has \$5:
\begin{itemize}
    \item With a consistent \$1 bet, $v_\pi(5) = 0.5$ indicates an expected gain of \$0.5.
    \item With a consistent \$2 bet, $v_\pi(5) = -1$ indicates an expected loss of \$1.
\end{itemize}

\textbf{Action Value function ($q_\pi(s, a)$)} assesses expected cumulative future rewards for action $a$ in state $s$. For instance:
\begin{itemize}
    \item $q_\pi(5, 1) = 1$ suggests a \$1 bet from \$5 results in a cumulative reward of \$1.
    \item $q_\pi(5, 2) = -0.5$ indicates a loss of \$0.5 for a \$2 bet from \$5.
\end{itemize}
This gambling game scenario highlights the role of state and action value functions in RL, guiding optimal decision-making in dynamic environments.


\subsection{Optimal Policies and Optimal Value Functions}\label{sec:OPOV}
Solving an RL task involves identifying a policy that maximizes long-term rewards. Value functions create a partial ordering over policies, allowing comparison and ranking based on expected cumulative rewards. A policy $\pi$ is better than or equal to $\pi_0$ if $v_\pi(s) \geq v_{\pi_0}(s)$ for all states $s$. An optimal policy is better than or equal to all other policies, denoted by $\pi_*$, sharing the same optimal state-value function $v_*$, which is defined as the maximum value function over all possible policies.
\begin{equation}
v_*(s) \equiv \max_\pi v_\pi(s) \quad \text{for all } s \in \mathcal{S}
\end{equation}
Optimal policies also share the same optimal action-value function $q_*$, which is defined as the maximum action value function over all possible policies.
\begin{equation}
q_*(s, a) \equiv \max_\pi q_\pi(s, a) \quad \text{for all } s \in \mathcal{S}
\end{equation}
The relationship between the optimal action value function \( q_*(s, a) \) and the optimal state value function \( v_*(s) \) is given by the following equation:
By having optimal action value function \( q_*(s, a) \) we can find optimal state value function \( v_*(s) \) as shown in Equation~\ref{Eq_VQ}.
\begin{equation}
\label{Eq_VQ}
q_*(s, a) = \mathbb{E}[\mathcal{R}_{t+1} + \gamma v_*(\mathcal{S}_{t+1}) \mid \mathcal{S}_t = s, \mathcal{A}_t = a]
\end{equation}

Optimal value functions and policies represent an ideal state in RL. It is however rare to find truly optimal policies in computationally demanding tasks due to practical challenges \cite{sutton2018reinforcement}. RL agents strive to approximate optimal policies. Dynamic Programming (DP) helps identify optimal values, assuming a perfect model of the environment, which is a challenge to have in real-world cases. DP methods are not also sample efficient, even though they are theoretically sound.
The fundamental idea of DP and RL is using value functions to organize the search for good policies. For finite MDPs, the environment's dynamics are given by probabilities $p(s', r \mid s, a)$.
The Bellman optimality equations for the optimal state value function \( v_*(s) \) and the optimal action value function \( q_*(s, a) \) are Equations~\ref{Eq_BOV} and ~\ref{Eq_BOQ}, respectively:

\begin{equation}
\label{Eq_BOV}
v_*(s) = \max_a \mathbb{E}[\mathcal{R}_{t+1} + \gamma v_*(\mathcal{S}_{t+1}) \mid \mathcal{S}_t = s, \mathcal{A}_t = a]
= \max_a \sum_{s', r} p(s', r \mid s, a) [r + \gamma v_*(s')]
\end{equation}

\begin{equation}
\label{Eq_BOQ}
q_*(s, a) = \mathbb{E}[\mathcal{R}_{t+1} + \max_{a'} q_*(\mathcal{S}_{t+1}, a') \mid \mathcal{S}_t = s, \mathcal{A}_t = a]
= \sum_{s', r} p(s', r \mid s, a) [r + \gamma \max_{a'} q_*(s', a')]
\end{equation}
DP algorithms are derived by transforming Bellman equations into update rules.

\subsection{Policy Evaluation (Prediction)}\label{sec:PE}

Policy evaluation, also known as prediction, involves computing the state-value function \( v_\pi \) for a given policy \( \pi \). This process assesses the expected return when following policy \( \pi \) from each state. The state-value function \( v_\pi(s) \) is defined as the expected return starting from state \( s \) and following policy \( \pi \):

\begin{equation}
v_\pi(s) \equiv \mathbb{E}_\pi [\mathcal{R}_{t+1} + \gamma \mathcal{G}_{t+1} \mid \mathcal{S}_t = s]
\end{equation}
This can be recursively expressed as:
\begin{equation}
v_\pi(s) = \mathbb{E}_\pi [\mathcal{R}_{t+1} + \gamma v_\pi(\mathcal{S}_{t+1}) \mid \mathcal{S}_t = s]
= \sum_a \pi(a \mid s) \sum_{s', r} p(s', r \mid s, a) [r + \gamma v_\pi(s')]
\end{equation}

In these equations, $\pi(a \mid s)$ denotes the probability of taking action $a$ in state $s$ under policy \( \pi \). The existence and uniqueness of \( v_\pi \) are guaranteed if \( \gamma < 1 \) or if all states eventually terminate under \( \pi \). Dynamic Programming (DP) algorithm updates are termed \textit{"expected updates"} because they rely on the expectation over all potential next states, rather than just a sample \cite{sutton2018reinforcement}.

\subsection{Policy Improvement}\label{sec:PI}
The purpose of calculating the value function for a policy is to identify improved policies. Assuming $v_\pi$ for a deterministic policy $\pi$, for a state $s$, should we alter the policy to select action $a \neq \pi(s)$? We know the effectiveness of adhering to the existing policy from state $s$ ($v_\pi(s)$), but would transition to a new policy yield a superior outcome? We can answer this by selecting action $a$ in $s$ and then following $\pi$. To determine if a policy can be improved, we compare the value of taking a different action \( a \) in state \( s \) with the current policy. This is done using the action value function \( q_\pi(s, a) \):
\begin{equation}
q_\pi(s, a) \equiv \mathbb{E}[\mathcal{R}_{t+1} + \gamma v_\pi(\mathcal{S}_{t+1}) \mid \mathcal{S}_t = s, \mathcal{A}_t = a]
= \sum_{s', r} p(s', r \mid s, a) [r + \gamma v_\pi(s')]
\end{equation}
The key criterion is whether this value exceeds $v_\pi(s)$. If $q_\pi(s, a) > v_\pi(s)$, consistently choosing action $a$ in $s$ is more advantageous than following $\pi$, leading to an improved policy $\pi'$.

\subsection{Policy Improvement Theorem}
The policy improvement theorem states that if \( q_\pi(s, \pi'(s)) \geq v_\pi(s) \) for all states \( s \), then the new policy \( \pi' \) is at least as good as the original policy \( \pi \). Formally, it is expressed as
\begin{equation}
q_\pi(s, \pi'(s)) \geq v_\pi(s)
\end{equation}
If $\pi'$ achieves greater or equal expected return from all states $s \in \mathcal{S}$:
\begin{equation}
v_{\pi'}(s) \geq v_\pi(s)
\end{equation}
If there is strict inequality at any state, $\pi'$ is superior to $\pi$. Extending this to all states and actions, selecting the action that maximizes $q_\pi(s, a)$. The new policy \( \pi' \) is obtained by selecting the action that maximizes the action value function \( q_\pi(s, a) \).
\begin{equation}
\pi'(s) \equiv \arg\max_a q_\pi(s, a)
= \arg\max_a \mathbb{E}[\mathcal{R}_{t+1} + \gamma v_\pi(\mathcal{S}_{t+1}) \mid \mathcal{S}_t = s, \mathcal{A}_t = a]
= \arg\max_a \sum_{s', r} p(s', r \mid s, a) [r + \gamma v_\pi(s')]
\end{equation}
Policy improvement creates a new policy that enhances an initial policy by adopting a greedy approach based on the value function. Assuming $\pi'$ is equally effective as $\pi$ but not superior, then $v_\pi = v_{\pi'}$ ensures that for all states $s \in \mathcal{S}$. The relationship between the optimal state value function \( v_*(s) \) and the optimal action value function \( q_*(s, a) \) is given by the following equation:

\begin{equation}
v_{\pi'}(s) = \max_a \mathbb{E}[\mathcal{R}_{t+1} + \gamma v_{\pi'}(\mathcal{S}_{t+1}) \mid \mathcal{S}_t = s, \mathcal{A}_t = a]
= \max_a \sum_{s', r} p(s', r \mid s, a) [r + \gamma v_{\pi'}(s')]
\end{equation}
Policy improvement yields a superior policy unless the initial policy is already optimal. This concept extends to stochastic policies. Stochastic policies introduce a set of probabilities for actions, with the action most aligned with the greedy policy assigned the highest probability.


\subsection{Policy Iteration}\label{sec:PIT}
After enhancing a policy $\pi$ using $v_\pi$ to derive an improved policy $\pi'$, compute $v_{\pi'}$ and further refine it to obtain a superior policy $\pi''$. This process generates a sequence of improving policies and corresponding value functions. The process of policy iteration involves alternating between policy evaluation and policy improvement to obtain a sequence of improving policies and value functions:

\begin{equation}
\pi_0 \xrightarrow{\text{Evaluation}} v_{\pi_0} \xrightarrow{\text{Improvement}} \pi_1 \xrightarrow{\text{Evaluation}} v_{\pi_1} \xrightarrow{\text{Improvement}} \pi_2 \xrightarrow{\text{Evaluation}} \ldots \xrightarrow{\text{Improvement}} \pi_* \xrightarrow{\text{Evaluation}} v_*
\end{equation}
Each policy in this sequence is a marked improvement over its predecessor unless the preceding one is already optimal. Given a finite MDP, this iterative process converges to an optimal policy and value function in a finite number of iterations. This method is called Policy Iteration. Policy iteration involves two processes: policy evaluation aligns the value function with the current policy, and policy improvement makes the policy greedier based on the value function. These processes iteratively reinforce each other until an optimal policy is obtained.

\subsection{Value Iteration}\label{sec:VI}
One limitation of policy iteration is that each iteration requires policy evaluation, often necessitating multiple passes through the entire state set \cite{bertsekas2011approximate}. To address this, policy evaluation can be abbreviated without losing convergence guarantees. This method, known as value iteration, terminates policy evaluation after a single sweep. It combines policy improvement with a truncated form of policy evaluation. Value iteration merges one pass of policy evaluation with policy improvement in each iteration, ensuring convergence to an optimal policy for discounted finite MDPs \cite{lutter2021value}. The update rule for value iteration is given in Equation~\ref{Eq_VI}.
\begin{equation}
\label{Eq_VI}
v_{k+1}(s) \equiv \max_a \mathbb{E}[\mathcal{R}_{t+1} + \gamma v_k(\mathcal{S}_{t+1}) \mid \mathcal{S}_t = s, \mathcal{A}_t = a]
= \max_a \sum_{s', r} p(s', r \mid s, a)[r + \gamma v_k(s')]
\end{equation}
In value iteration, the key advantage is its efficiency, as it reduces the computational burden by merging policy evaluation and improvement into a single update step. This method is particularly useful for large state spaces where full policy evaluation at each step of policy iteration is computationally prohibitive \cite{sutton2018reinforcement}. Additionally, value iteration can be implemented using a synchronous update approach, where all state values are updated simultaneously, or an asynchronous update approach, where state values are updated one at a time, potentially allowing for faster convergence in practice. Another notable aspect of value iteration is its robustness to initial conditions. Starting from an arbitrary value function, value iteration iteratively refines the value estimates until convergence, making it a reliable method for finding optimal policies even when the initial policy is far from optimal \cite{bertsekas2012dynamic}. Furthermore, value iteration provides a foundation for more advanced algorithms by illustrating the principle of bootstrapping, where the value of a state is updated based on the estimated values of successor states. This principle is central to many RL algorithms that seek to balance exploration and exploitation in dynamic and uncertain environments \cite{mnih2015human}.

The concepts introduced in this section—ranging from bandits and MDPs to value functions and policies—provide the mathematical and conceptual tools required to understand RL methods. In the next section, we explore how these foundational ideas evolve into core RL algorithms, bridging theory and application.

\section{Core RL Methods}\label{sec:Terminology}
Understanding the various methodologies and concepts within RL is essential for the effective design and implementation of RL algorithms. Methods in RL can be classified as either off-policy or on-policy, and as model-free and model-based. These categories offer different approaches and techniques for learning from interactions with the environment.

\subsection{Model-free \& Model-based methods}
\textbf{Model-free} methods determine the optimal policy or value function directly without constructing a model of the environment. There is no requirement for them to know transition probabilities and rewards, as they learn entirely from observed states, actions, and rewards. Compared with model-based methods, model-free methods are simpler to implement, relying on experience-based learning. There are two primary types: \textit{Value-based} and \textit{Policy-based} methods. The former focus on learning the action-value function to derive an optimal policy. For instance, Q-learning (discussed in section \ref{sec:Algs}) is an off-policy algorithm that learns the value of the optimal policy independently of the agent’s actions by using a max operator in its update rule. SARSA (also discussed in section \ref{sec:Algs}), on the other hand, is an on-policy algorithm that updates its Q-values based on the actions actually taken by the policy. Both methods update their action-value estimates based on the Bellman equation until convergence. In contrast, policy-based methods, like REINFORCE (discussed in section \ref{sec:Algs}), work by directly learning the policy without explicitly learning a value function. These methods adjust the policy parameters directly by following the gradient of the expected reward. This approach is particularly useful in environments with high-dimensional action spaces where value-based methods may not be effective. Policy-based methods are also capable of handling stochastic policies, providing a natural framework for dealing with uncertainty in action selection.
In addition to these primary types, there are also \textbf{hybrid approaches} that combine value-based and policy-based methods, such as Actor-Critic algorithms (which will be discussed in section \ref{sec:Algs}). These methods consist of two main components: an \textbf{actor} that updates the policy parameters in a direction suggested by the critic, and a \textbf{critic} that evaluates the action-value function. Combining both types of learning is intended to provide more stable and efficient learning \cite{grondman2012survey}.

Another significant advancement in model-free methods is the development of Deep RL (DRL) By integrating deep neural networks with traditional RL algorithms, methods such as Deep Q-Networks (DQN) \cite{mnih2013playing} and Proximal Policy Optimization (PPO) \cite{schulman2017proximal} have achieved remarkable success in complex, high-dimensional environments, including games and robotic control tasks. The advancement of these technologies has opened up new possibilities for the application of RL to real-world problems, enabling the demonstration of robust performance in domains which were previously intractable. It is beyond the scope of this paper to discuss these algorithms, and wee refer you to \cite{li2017deep, arulkumaran2017brief, franccois2018introduction, li2023deep} to understand DRL deeply and effectively.

It is possible to predict the outcomes of actions using \textbf{model-based} methods, which facilitate strategic planning and decision-making. The use of these methods enhances learning efficiency by providing opportunities for virtual experimentation, despite the complexity of developing and refining accurate models \cite{luo2024survey}. Autonomous driving systems are an example of how model-based methods can be applied in the real world. As autonomous vehicles navigate in dynamic environments, obstacle avoidance, and optimal routing must be made in real time.
Autonomous vehicles create detailed models of their environment. These models include static elements, such as roads and buildings, as well as dynamic elements, such as other vehicles and pedestrians. Sensor data, including cameras, LIDAR, and radar, are used to build this model.
Through the use of the environmental model, the vehicle is capable of predicting the outcome of various actions. For instance, when a vehicle considers changing lanes, it uses its model to predict the behavior of surrounding vehicles to determine the safest and most efficient way to make the change.
The model assists the vehicle in planning its route and making strategic decisions. To minimize travel time, avoid congestion, and enhance safety, it evaluates different routes and actions. Simulation allows the vehicle to select the best course of action by simulating various scenarios before implementation in the real world.
The vehicle, for example, may use the model to simulate different actions in the event of a busy intersection, such as waiting for a gap in traffic or taking an alternate route. Considering the potential outcomes of each action, the vehicle can make an informed decision that balances efficiency with safety. In addition to improving the ability of autonomous vehicles to navigate safely and efficiently in real-world conditions, this model-based approach enables them to make complex decisions with a high level of accuracy. As a result of continuously refining the model based on new data, the vehicle is able to enhance its decision-making capabilities over time, thereby improving performance and enhancing safety on the road.

There are several advantages to using model-based methods over methods that do not use models. By simulating future states and rewards, they can plan and evaluate different action sequences without interacting directly with the environment. It is believed that this capability may lead to a faster convergence to an optimal policy, since learning can be accelerated by leveraging the model's predictions. A model-based approach can also adapt more quickly to changes in the environment, since it enables the model to be updated and re-planned accordingly.
Although model-based methods have many advantages, they also face a number of challenges, primarily in regards to accuracy and computational cost. In order to create an accurate model of the environment, a high-fidelity model needs to be created. Moreover, the planning process may be computationally expensive, especially in environments with a large number of states and actions. However, advances in computing power and algorithms continue to improve the feasibility and performance of model-based methods, making them a valuable approach in RL \cite{moerland2023model}.

\subsection{Off-Policy and On-Policy Methods}

On-policy and off-policy learning are methodologies within model-free learning approaches, not relying on environment transition probabilities. They are classified based on the relationship between the behavior policy and the updated policy \cite{sutton2018reinforcement}.
\textbf{On-policy} methods evaluate and improve the policy used to make decisions, intertwining exploration and learning. These methods update the policy based on the actions taken and the rewards received while following the current policy ($\pi$). This ensures that the policy being optimized is the one actually used to interact with the environment, allowing for a coherent learning process where exploration and policy improvement are naturally integrated.

\textbf{Off-policy} methods, on the other hand, involve learning the value of the optimal policy independently of the agent's actions. In these methods, we distinguish between two types of policies: the \textit{behavior policy} ($b$) and the \textit{target policy} ($\pi$). The behavior policy explores the environment, while the target policy aims to improve performance based on the gathered experience. This allows for a more exploratory behavior policy while learning an optimal target policy. A significant advantage of off-policy methods is that they can learn from data generated by any policy, not just the one currently being followed, making them highly flexible and sample-efficient.
The decoupling of the behavior and target policies allows off-policy methods to reuse experiences more effectively. For instance, experiences collected using a behavior policy that explores the environment broadly can be used to improve the target policy, which aims to maximize rewards. This characteristic makes off-policy methods particularly powerful in dynamic and complex environments where extensive exploration is required \cite{dayan1992q, hessel2018rainbow}.

The relationship between the target policy and the behavior policy determines if a method is on-policy or off-policy. Identical policies indicate on-policy, while differing policies indicate off-policy. Implementation details and objectives also influence classification. To better distinguish these methods, we have to first learn what are the different policies.
\textbf{Behavior Policy $b$} is a strategy used by an agent to determine which actions to take at each time step. The behavior policy might, for example, include recommending a variety of movies in order to explore user preferences in the recommendation system example.
\textbf{Target policy $\pi$} governs how the agent updates its value estimates in response to observed outcomes. Depending on the feedback received from the recommended movies, the target policy of the recommendation system may update the estimated user preferences.A thorough understanding of the interactions between these policies is essential for the implementation of effective learning systems. An agent's behavior policy determines how it explores an environment, balancing exploration with exploitation to gather useful information. Alternatively, the target policy determines how the agent learns from these experiences in order to improve its estimates of value.
When using on-policy methods, the behavior policy and the target policy are the same, meaning that the actions taken to interact with the environment are also used to update the value estimates. The result is stable learning, but it can be less efficient because the policy may not sufficiently explore the state space \cite{rummery1994line}.
There is a difference between the behavior policy and the target policy in off-policy methods. As opposed to the behavior policy, the target policy focuses on optimizing the value estimates by taking the most appropriate action. Despite the fact that this separation can make learning more efficient, it can also introduce instability if the behavior policy diverges too far from the optimal policy \cite{mnih2015human, hessel2018rainbow}.
Furthermore, advanced methods, such as Actor-Critic algorithms, separate the \textit{behavior policy (actor)} and the \textit{target policy (critic)}. Actors make decisions according to current policies, while critics evaluate these decisions and provide feedback to improve policies, thus combining the stability of on-policy methods with the efficiency of off-policy methods \cite{konda1999actor, lillicrap2015continuous}.

Understanding the core methodologies of RL, such as model-free and model-based approaches, as well as the distinction between off-policy and on-policy methods, provides a foundational framework for exploring the diverse landscape of RL algorithms. These methodologies not only shape how agents learn from their environments but also influence their adaptability and efficiency in complex, real-world scenarios.
Building on this understanding, the next section delves deeper into the specific essential algorithms underpinning these methods, focusing on the policy-based, value-based, and hybrid approaches, to provide a clearer picture of their mechanisms and applications in RL.

\section{Essential Algorithms}\label{sec:Algs}
This section aims to present a concise overview of key algorithms discussed thus far, accompanied by references to the original research papers for further exploration. Each algorithm is briefly described, and real-world examples are included to enhance understanding. Readers seeking detailed information are encouraged to consult the cited references, which serve as a gateway to the primary sources and support a deeper learning experience. We categorize algorithms into three types: Value-based, Policy-based, and Hybrid Algorithms. For each type, we analyze one to two widely-used algorithms, acknowledging that there are more algorithms to discover.

\subsection{Value-based}
We introduced value-based methods, and clearly analyzed how they work. Here, we introduce three algorithms, that are in the tabular settings. Later, we dive deeper into the topics, and discuss value-based methods that use Deep Learning, such as Deep Q-Networks.

A significant breakthrough was made by \cite{watkins1989learning} with the introduction of \textbf{Q-learning}, a Model-free algorithm considered as off-policy Temporal Difference (TD) control. TD learning is undoubtedly the most fundamental and innovative concept. A combination of Monte Carlo (MC) methods and Dynamic Programming (DP) is used in this method. On one hand, similar to MC approaches, TD learning can be used to acquire knowledge from unprocessed experience without the need for a model that describes the dynamics of the environment. On the other hand, TD algorithms are also similar to DP in that they refine predictions using previously learned estimates instead of requiring a definitive outcome in order to proceed (known as \textit{bootstrapping}).
Q-learning enables an agent to learn the value of an action in a particular state through experience, without requiring a model of the environment. It operates on the principle of learning an action-value function that gives the expected utility of taking a given action in each state and following a fixed policy thereafter. 
The core of the Q-learning algorithm involves updating the Q-values (action-value pairs), where the learned action-value function, denoted as \( \mathcal{Q} \), approximates \( q_* \), the optimal action-value function, regardless of the policy being followed. This significantly simplifies the algorithm's analysis and has facilitated early proofs of convergence. However, the policy still influences the process by determining which state-action pairs are visited and subsequently updated.

\begin{equation}
\begin{aligned}
\mathcal{Q}(\mathcal{S}_t, \mathcal{A}_t) \leftarrow \mathcal{Q}(\mathcal{S}_t, \mathcal{A}_t) 
+ \alpha \big[\mathcal{R}_{t+1} + \gamma \max_a \mathcal{Q}(\mathcal{S}_{t+1}, a) - \mathcal{Q}(\mathcal{S}_t, \mathcal{A}_t) \big]
\end{aligned}
\end{equation}

There are other types of Q-learning introduced in the literature, with slight changes and improvements, such as: Double Q-learning \cite{hasselt2010double}, that addresses the overestimation bias in Q-learning, Distributional Q-learning \cite{bellemare2017distributional}, which models the distribution of returns instead of estimating the mean Q-value, providing richer information for decision-making, and many more \cite{matignon2007hysteretic, haarnoja2018soft, wang2020qplex}.

Another widely used value-based algorithm is \textbf{Deep Q-Networks (DQN)}, which merges \textit{Q-learning} with Neural Networks to learn control policies directly from raw pixel inputs. It uses Convolutional Neural Networks (CNN) to process these inputs and an experience replay mechanism to stabilize learning by breaking correlations between consecutive experiences. The target network, updated less frequently, aids in stabilizing training. DQN achieved state-of-the-art performance on various Atari 2600 games, surpassing previous methods and, in some cases, human experts, using a consistent network architecture and hyperparameters across different games \cite{mnih2013playing}.
DQN combines the introduced Bellman Equation with DL approaches like Loss Function and Gradient Descent to find the optimal policy as below:

\begin{equation}
\mathcal{L}_i(\theta_i) = \mathbb{E}_{(s,a,r,s') \sim \mathcal{D}} \left[ \left( y_i - \mathcal{Q}(s, a; \theta_i) \right)^2 \right]
\end{equation}
where
\begin{equation}
y_i = r + \gamma \max_{a'} \mathcal{Q}(s', a'; \theta^-)
\end{equation}

\begin{equation}
\begin{aligned}
\nabla_{\theta_i} \mathcal{L}_i(\theta_i) = \mathbb{E}_{(s,a,r,s') \sim \mathcal{D}} \Big[ & \left( r + \gamma \max_{a'} \mathcal{Q}(s', a'; \theta^- ) \right. \left. - \mathcal{Q}(s, a; \theta_i) \right) \nabla_{\theta_i} \mathcal{Q}(s, a; \theta_i) \Big]
\end{aligned}
\end{equation}

Similar to Q-learning, there have been updates made to DQN. Some of the variations are \cite{van2016deep, wang2016dueling, schaul2015prioritized}.

\begin{table}[t]
\centering
\caption{Essential RL Algorithms}
\small\addtolength{\tabcolsep}{-5 pt}
\begin{tabular}{|c|p{10cm}|}
    \hline
    \rowcolor[gray]{0.45} \multicolumn{2}{|c|}{\textbf{Algorithms}} \\ \hline
    1 & \textbf{Q-Learning \cite{watkins1989learning}} - Model-free, Off-policy, Value-based \\ \hline
    2 & \textbf{SARSA (State-Action-Reward-State-Action) \cite{rummery1994line}} - Model-free, On-policy, Value-based \\ \hline
    3 & \textbf{Expected SARSA \cite{van2009theoretical}} - Model-free, On-policy, Value-based \\ \hline
    4 & \textbf{REINFORCE \cite{williams1992simple}} - Model-free, On-policy, Policy-based \\ \hline
    5 & \textbf{Dyna-Q \cite{sutton1990integrated}} - Model-based, Off-policy, Hybrid \\ \hline
    6 & \textbf{DQN} \cite{mnih2013playing} - Model-free, Off-policy, Value-based \\ \hline
    7 & \textbf{TRPO} \cite{schulman2015trust} - Model-free, On-policy, Policy-based \\ \hline
    8 & \textbf{PPO} \cite{schulman2017proximal} - Model-free, On-policy, Policy-based \\ \hline
    9 & \textbf{SAC (Soft Actor-Critic) \cite{haarnoja2018soft}} - Model-free, Off-policy, Hybrid \\ \hline
    10 & \textbf{A3C} \cite{mnih2016asynchronous} - Model-free, On-policy, Hybrid \\ \hline
    11 & \textbf{A2C} \cite{lillicrap2015continuous} - Model-free, On-policy, Hybrid \\ \hline
    12 & \textbf{DDPG (Deep Deterministic Policy Gradient) \cite{lillicrap2015continuous}} - Model-free, Off-policy, Policy-based \\ \hline
    13 & \textbf{TD3 (Twin Delayed Deep Deterministic Policy Gradient) \cite{fujimoto2018addressing}} - Model-free, Off-policy, Policy-based \\ \hline
\end{tabular}
\label{tab:rl_algorithms}
\end{table}

\subsection{Policy-based}
Moving on from value-based methods, we analyze some policy-based algorithms in this section. \textbf{Policy-based} methods are another fundamental RL method that more strongly emphasizes direct policy optimization in the process of choosing actions for an agent. In contrast to Value-based methods, which search for the value function implicit in the task, and then derive an optimal policy, Policy-based methods directly parameterize and optimize the policy. This approach offers several advantages, particularly better dealing with very challenging environments that have high-dimensional action spaces or where policies are inherently stochastic. Perhaps at the core, Policy-based methods conduct their operation based on the parameterization of policies, usually denoted as $\pi(a|s;\theta)$. Here, $\theta$ is used to denote the parameters of the policy, while $s$ denotes the state and $a$ denotes the action. In other words, it finds the optimal parameters $\theta^*$ that maximize the expected cumulative reward. Needless to say, this is generally done by gradient ascent techniques and more specifically by Policy Gradient methods that explicitly compute the gradient of expected reward with respect to the policy parameters, modifying parameters in the direction of reward increase \cite{sutton2018reinforcement, kaelbling1996reinforcement, kober2013reinforcement}.

\textbf{REINFORCE} is one of the widely-used policy-based algorithms. The REINFORCE algorithm is a seminal contribution to RL, particularly within the context of policy gradient methods. The algorithm is designed to optimize the expected cumulative reward by adjusting the policy parameters in the direction of the gradient of the expected reward. It is rooted in the stochastic policy framework, where the policy, parameterized by \(\theta\), defines a probability distribution over actions given the current state. The key insight of the REINFORCE algorithm is to use the log-likelihood gradient estimator to update the policy parameters \cite{williams1992simple}. The gradient of the expected reward with respect to the policy parameters \(\theta\) is given by:

\begin{equation}
\nabla_\theta \mathcal{J}(\theta) = \mathbb{E}_\pi \left[ \nabla_\theta \log \pi_\theta(a|s) \mathcal{G}_t \right],
\end{equation}
where \(\pi_\theta(a|s)\) is the probability of taking action \(a\) in the state \(s\) under policy \(\pi\) parameterized by \(\theta\), and \(\mathcal{G}_t\) is the return (cumulative future reward) following time step \(t\). This gradient estimation forms the basis for the parameter update rule.
\begin{equation}
\theta \leftarrow \theta + \alpha \nabla_\theta \log \pi_\theta(a|s) \mathcal{G}_t,
\end{equation}
where \(\alpha\) is the learning rate. 

Another algorithm, which has been used in variety of applications \cite{wei2019mixed, zhang2020real, jin2021optimal, zhang2021image, guan2020centralized, bohn2019deep, lopes2018intelligent, ye2020automated}, is \textbf{Proximal Policy Optimization (PPO)}. PPO, proposed by \cite{schulman2017proximal}, represents a significant advancement within policy gradient methods. PPO aims to achieve reliable performance and sample efficiency, addressing the limitations of previous policy optimization algorithms such as Vanilla Policy Gradient (VPG) \cite{sutton1999policy} methods and Trust Region Policy Optimization (TRPO) \cite{schulman2015trust}. 
Using policy gradient methods, the policy parameters are optimized through stochastic gradient ascent by estimating the gradient of the policy. One of the most commonly used policy gradient estimators is:

\begin{equation}
    \hat{g} = \hat{\mathbb{E}}_t \left[ \nabla_\theta \log \pi_\theta(a_t | s_t) \hat{A}_t \right],
\end{equation}
where \(\pi_\theta\) represents the policy parameterized by \(\theta\), and \(\hat{A}_t\) is an estimator of the advantage function at time step \(t\). This estimator helps construct an objective function whose gradient corresponds to the policy gradient estimator:
\begin{equation}
    \mathcal{L}_{\mathcal{PG}}(\theta) = \hat{\mathbb{E}}_t \left[ \log \pi_\theta(a_t | s_t) \hat{A}_t \right].
\end{equation}
PPO simplifies TRPO by using a surrogate objective with a clipped probability ratio, allowing for multiple epochs of mini-batch updates. In order to preserve learning, large policy updates should be avoided.

\subsection{Hybrid (Actor-Critic) methods}
For the last group of algorithm, hybrid methods, we introduce \textbf{Asynchronous Advantage Actor-Critic (A3C)} and \textbf{Advantage Actor-Critic (A2C)}. Actor-critic methods combine Value-based and Policy-based approaches. Essentially, these methods consist of two components: the Actor, who selects actions based on a policy, and the Critic, who evaluates the actions based on their value function. By providing feedback on the quality of the actions taken, the critic guides the actor in updating the policy directly. As a result of this synergy, learning can be more stable and efficient, addressing some limitations of pure policy or Value-based approaches \cite{grondman2012survey, arulkumaran2017deep}.
The A2C algorithm is a synchronous variant of the A3C algorithm, which was introduced by \cite{mnih2016asynchronous}. A2C  maintains the key principles of A3C but simplifies the training process by synchronizing the updates of multiple agents, thereby leveraging the strengths of both Actor-Critic methods and advantage estimation.
The Actor-Critic architecture combines two primary components, in both algorithms: the actor, which is responsible for selecting actions, and the critic, which evaluates the actions by estimating the value function. The actor updates the policy parameters in a direction that is expected to increase the expected reward, while the critic provides feedback by computing the TD error. This integration allows for more stable and efficient learning compared to using Actor-only or critic-only methods \cite{konda1999actor}.
Advantage estimation is a technique used to reduce the variance of the policy gradient updates. The advantage function $\mathcal{A}(s, a)$ represents the difference between the action-value function $\mathcal{Q}(s, a)$ and the value function $\mathcal{V}(s)$.

\begin{equation}
    \mathcal{A}(s, a) = \mathcal{Q}(s, a) - \mathcal{V}(s).
\end{equation}

By using the advantage function, A2C focuses on actions that yield higher returns than the average, which helps in making more informed updates to the policy \cite{sutton2018reinforcement}.
Unlike A3C, where multiple agents update the global model asynchronously, A2C synchronizes these updates. Multiple agents run in parallel environments, collecting experiences and calculating gradients, which are then aggregated and used to update the global model synchronously. This synchronization reduces the complexity of implementation and avoids issues related to asynchronous updates, such as non-deterministic behavior and potential overwriting of gradients. Table \ref{tab:rl_algorithms} categorizes the examined algorithms, and other essential algorithms that we did not discuss to give a comprehensive overview regarding the main features

\begin{table}[t]
    \centering
    \caption{Reinforcement Learning Resources}
    \label{tab:rl_resources}
    \small\addtolength{\tabcolsep}{-5 pt}
    \begin{tabular}{|c|p{10cm}|}
        \hline
        \rowcolor[gray]{0.45} \multicolumn{2}{|c|}{\textbf{Books}} \\ \hline
        1 & \textbf{"Reinforcement Learning: An Introduction"} by Richard S. Sutton and Andrew G. Barto \\ \hline
        2 & \textbf{"Deep Reinforcement Learning Hands-On"} by Maxim Lapan \\ \hline
        3 & \textbf{"Grokking Deep Reinforcement Learning"} by Miguel Morales \\ \hline
        4 & \textbf{"Algorithms for Reinforcement Learning"} by Csaba Szepesvári \\ \hline
        
        \rowcolor[gray]{0.45} \multicolumn{2}{|c|}{\textbf{Online Courses}} \\ \hline
        5 & \textbf{Coursera: Reinforcement Learning Specialization} by University of Alberta \\ \hline
        6 & \textbf{Udacity: Deep Reinforcement Learning Nanodegree} \\ \hline
        7 & \textbf{edX: Fundamentals of Reinforcement Learning} by University of Alberta \\ \hline
        8 & \textbf{Reinforcement Learning Winter 2019} (Stanford) \\ \hline

        \rowcolor[gray]{0.45} \multicolumn{2}{|c|}{\textbf{Video Lectures}} \\ \hline
        9 & \textbf{\href{https://www.youtube.com/playlist?list=PLqYmG7hTraZCDxZ44o4p3N5Anz3lLRVZF}{DeepMind x UCL | Reinforcement Learning Lecture Series}} \\ \hline
        10 & \textbf{\href{https://www.youtube.com/playlist?list=PLqYmG7hTraZBKeNJ-JE_eyJHZ7XgBoAyb}{David Silver's Reinforcement Learning Course}} \\ \hline
        11 & \textbf{\href{https://www.youtube.com/playlist?list=PLdAoL1zKcqTXFJniO3Tqqn6xMBBL07EDc}{Pascal Poupart's Reinforcement Learning Course - CS885}} \\ \hline
        12 & \textbf{\href{https://www.youtube.com/playlist?list=PLImtCgowF_ES_JdF_UcM60EXTcGZg67Ua}{Sarath Chandar's Reinforcement Learning Course}} \\ \hline

        \rowcolor[gray]{0.45} \multicolumn{2}{|c|}{\textbf{Tutorials and Articles}} \\ \hline
        13 & \textbf{\href{https://spinningup.openai.com/en/latest/}{OpenAI Spinning Up in Deep RL}} \\ \hline
        14 & \textbf{\href{https://pytorch.org/tutorials/intermediate/reinforcement_q_learning.html}{Deep Reinforcement Learning Course by PyTorch}} \\ \hline
        15 & \textbf{\href{https://github.com/dennybritz/reinforcement-learning}{RL Adventure by Denny Britz}} \\ \hline

        \rowcolor[gray]{0.45} \multicolumn{2}{|c|}{\textbf{Online Communities and Forums}} \\ \hline
        16 & \textbf{\href{https://www.reddit.com/r/reinforcementlearning/}{Reddit: r/reinforcementlearning}} \\ \hline
        17 & \textbf{\href{https://stackoverflow.com/questions/tagged/reinforcement-learning}{Stack Overflow}} \\ \hline
        18 & \textbf{\href{https://www.alignmentforum.org/}{AI Alignment Forum}} \\ \hline
    \end{tabular}
\end{table}

By analyzing some of the widely-used algorithms, it is time to introduce some of the good resources to further learn RL. In the next section, we introduce books, video lectures, and online communities in RL.
\section{Resources and Further Reading}\label{sec:Resources}
In this section, we provide a list of some of the most helpful books, courses, videos, and online communities to assist readers in getting started with their real-life research without being overwhelmed. Table~\ref{tab:rl_resources} summarizes the necessary resources, all in one place. On top of the mentioned resources, we would like to refer readers to survey papers \cite{kaelbling1996reinforcement, li2017deep, arulkumaran2017deep, wang2022deep, wang2020deep, moerland2023model, polydoros2017survey, luo2024survey, sato2019model, ramirez2022model} that are really helpful to understand different applications, algorithms, and background more.

\section{Conclusion}
\label{sec:conclusion}

This paper presents an introductory exploration of the fundamental concepts and methodologies of Reinforcement Learning (RL), tailored to beginners. It establishes a foundational understanding of how RL agents learn and make decisions by thoroughly examining key components such as states, actions, policies, and reward signals. By analyzing Multi-armed bandit problem, this paper introduced background of RL in an accessible and easy-to-understand way. The primary objective is to offer an overview of a wide range of RL algorithms, encompassing both model-free and model-based approaches, thereby highlighting the diversity within the field. Through this guide, we aim to equip new learners with the essential knowledge and confidence to begin their journey into RL.

\bibliographystyle{ieeetr}   
\bibliography{references}      

\end{document}